\pdfoutput=1

\documentclass[11pt]{article}

\usepackage[final]{acl}

\usepackage{times}
\usepackage{latexsym}

\usepackage[T1]{fontenc}

\usepackage[utf8]{inputenc}

\usepackage{microtype}

\usepackage{inconsolata}

\usepackage{natbib}
\usepackage{multirow}
\usepackage{algorithm}
\usepackage{algorithmic}
\usepackage{hyperref}       
\usepackage{url}            
\usepackage{booktabs}       
\usepackage{amsfonts}       
\usepackage{nicefrac}       
\usepackage{xcolor}         
\usepackage{color}

\usepackage{graphicx}
\usepackage{relsize}
\usepackage{amsfonts}
\usepackage{url}
\usepackage{graphicx}
\usepackage{subfigure}
\usepackage{caption}
\usepackage{hhline}
\usepackage{colortbl}
\usepackage{tabularx}
\usepackage{amsfonts}
\usepackage{amssymb}
\usepackage{amsthm,amsmath}
\usepackage{mathrsfs}
\usepackage{setspace}
\usepackage{lipsum}

\usepackage{algorithm,algorithmic}
\usepackage{multicol}
\definecolor{myred}{RGB}{222,110,102} 
\definecolor{myblue}{RGB}{80,150,222} 
\definecolor{myyellow}{RGB}{203, 222, 58} 

\title{MetaRM: Shifted Distributions Alignment via Meta-Learning}

\author{\textbf{Shihan Dou}$^{1}\thanks{{ }{ }Equal contribution.}$, 
\textbf{Yan Liu}$^{1 *}$, 
\textbf{Enyu Zhou}$^{1}$, 
\textbf{Songyang Gao}$^{1}$, 
\textbf{Tianlong Li}$^{1}$, 
\\
\textbf{Haoxiang Jia}$^{2}$, 
\textbf{Limao Xiong}$^{1}$\textbf{,} 
\textbf{Xin Zhao}$^{1}$\textbf{,}
\textbf{Junjie Ye}$^{1}$\textbf{,}
\\
\textbf{Rui Zheng}$^{1}$\textbf{,} 
\textbf{Tao Gui}$^{1}$\thanks{{ }{ }Corresponding author.}\textbf{\ ,} 
\textbf{Qi Zhang}$^{1}$\textbf{,} 
\textbf{Xuanjing Huang}$^{1}$
\\
$^{1}$ NLP Group, Fudan University\\
$^{2}$ Peking University \\
\texttt{shdou21@m.fudan.edu.cn}\\
\texttt{\{rzheng20, tgui, qz\}@fudan.edu.cn}
}

\begin{document}
\maketitle
\begin{abstract}

The success of Reinforcement Learning from Human Feedback (RLHF) in language model alignment is critically dependent on the capability of the reward model (RM). 
However, as the training process progresses, the output distribution of the policy model shifts, leading to the RM's reduced ability to distinguish between responses. 
This issue is further compounded when the RM, trained on a specific data distribution, struggles to generalize to examples outside of that distribution.
These two issues can be united as a challenge posed by the shifted distribution of the environment.
To surmount this challenge, we introduce MetaRM, a method leveraging meta-learning to align the RM with the shifted environment distribution.
MetaRM is designed to train the RM by minimizing data loss, particularly for data that can improve the differentiation ability to examples of the shifted target distribution.
Extensive experiments demonstrate that MetaRM significantly improves the RM's distinguishing ability in iterative RLHF optimization, and also provides the capacity to identify subtle differences in out-of-distribution samples \footnote{\ The code will be made available upon publication.}.


\end{abstract}

\begin{figure}[t]
\centering
\includegraphics[width=0.47\textwidth]{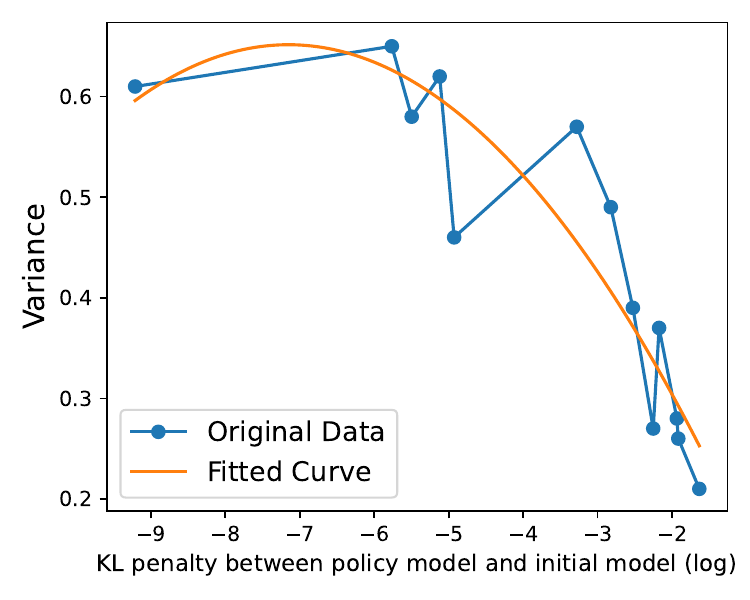}
\vspace{-0.5em}
\caption{
Variance of reward difference distribution.
We select 1K queries randomly and for each query, we sample two responses from the model output distribution and compute the difference between these rewards, to obtain the reward difference distribution.
As the RL training process progresses, the model output distribution shifts, causing the RM to fail to distinguish between responses, resulting in a decreasing variance.
These indicate that the RM struggles to capture subtle differences between responses under conditions of shifting environment distribution.
}
\vspace{-0.8em}
\label{fig: Reward Difference Distribution}
\end{figure}

\section{Introduction}
Reinforcement learning from human feedback (RLHF) provides a pivotal technique to ensure that the behavior of AI systems aligns with the intentions of their designers and the expectations of users \citep{DBLP:journals/corr/abs-2204-05862, ouyang2022training, zheng2023secrets}.
RLHF is executed in two primary stages. 
The initial stage involves training a reward model using preference data, which is collected from a substantial number of crowdsource workers. 
The second stage entails the application of reinforcement learning (RL) to fine-tune the large language model (LLM), to maximize the reward. 
In this process, the reward model plays a pivotal role, as its performance significantly impacts the effectiveness of the LLM's fine-tuning \cite{eschmann2021reward, gao2022scaling}.


However, researchers have pointed out that the reward model faces generalization challenges caused by the environment distribution shifts \cite{casper2023open, di2022goal}.
\textbf{Firstly}, as the RL training process progresses, the output distribution of the language model shifts, which leads the reward model to fail to distinguish between responses sampled from the same prompts, as shown in Figure~\ref{fig: Reward Difference Distribution}.
\textbf{Secondly}, the reward model trained on data from a specific distribution may struggle with out-of-distribution (OOD) examples in the RL training phase \cite{casper2023open, wulfe2022dynamics}.
Such limitations can lead to instability in the RL process.
Although \citet{touvron2023llama} proposes to iteratively collect preference pairs and fine-tune the reward model to maintain it in the new distribution, continuously collecting new data is resource and time-intensive.
The challenge of aligning the reward model with a new distribution when an environment distribution shift occurs has not been thoroughly examined.




In this paper, we introduce MetaRM, a novel approach that aligns the reward model with the new distribution through meta-learning to recover the reward model’s distinguishing ability.
The key insight of our method is that the reward model should minimize the loss of data, particularly those that can improve the differentiation ability to examples of the shifted target distribution.
In this way, we can bridge the gap between the preference data distribution and the model output distribution.
It ensures that the reward model not only performs well on the preference data but also can distinguish the differences in target domain outputs.
By using MetaRM, we can train new reward models to adapt to the output distribution of the newly aligned model, achieving iterative RLHF.
Additionally, our proposed approach also makes the reward model trained only on specific distribution preference data that can be applied to OOD data.

To evaluate the effectiveness of MetaRM, we apply it to the Anthropic's HH-RLHF \cite{DBLP:journals/corr/abs-2204-05862} and OpenAI's summarization \cite{DBLP:journals/corr/abs-2009-01325} datasets.
The experimental results demonstrate that our method can make the reward model restore the distinguishing ability in iterative RLHF optimization.
It can consistently achieve improvement of the language model in $3$ to $4$ rounds by iteratively training the reward model on original preference data.
In addition, we also evaluate MetaRM in an OOD setting and the results show that it also can maintain the ability to differentiate subtle differences in OOD samples.
The main contributions of our paper are as follows:
\begin{itemize}
\item We introduce MetaRM, a novelty method that makes the reward model adapt to the new environment distribution through meta-learning, which achieves to improve the language model by iterative RLHF.
\item MetaRM also enables the reward model trained only on specific distribution preference data that can be effectively applied to OOD data, without the need for laboriously labeling data on the target distribution.
\item Experiments show that MetaRM can make the reward model maintain the ability to differentiate between responses sampled from shifted distribution under the same prompts.
\end{itemize}


\begin{figure*}[t]
\centering
\includegraphics[width=0.96\textwidth]{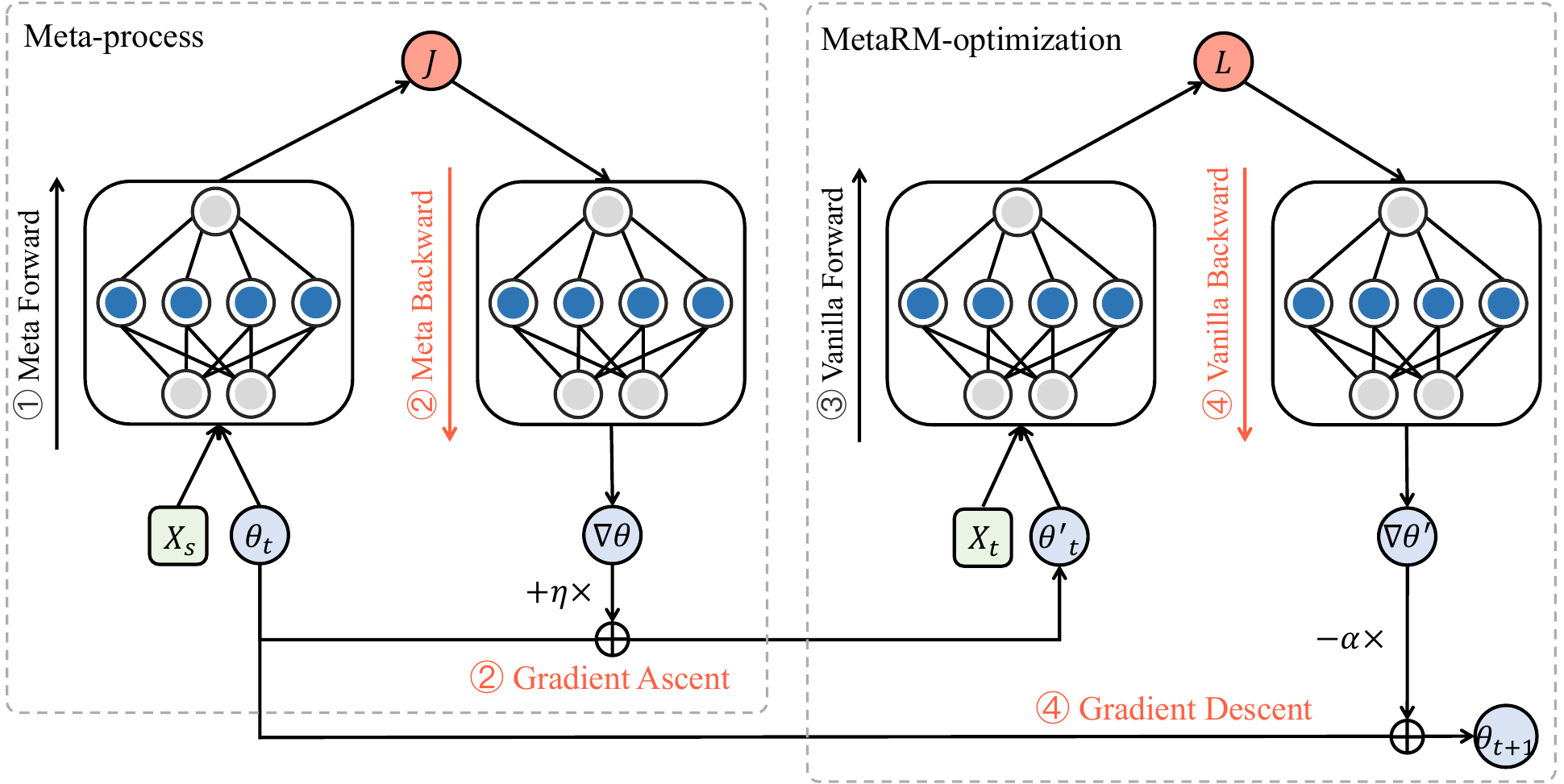}
\caption{The pipeline of our proposed approach MetaRM.
MetaRM contains four simple steps: 
1. Compute the difference loss on responses sampled from the shifted distribution.
2. Calculate the gradient of this loss wrt. the RM parameters $\theta_t$ and adjust the parameters according to the ascent direction.
3. Compute the vanilla loss on the original preference pairs using the updated parameters $\theta_{t}^{'}$.
4. Calculate the gradient of the vanilla loss wrt. $\theta_{t}^{'}$ and optimize the original parameters $\theta$ following the descent direction.
}
\label{fig: MetaRM Routine}
\vspace{-0.5em}
\end{figure*}

\section{Related Work}
\textbf{Reinforcement Learning from Human Feedback.}
Previous studies have demonstrated that RLHF \cite{DBLP:journals/corr/abs-2204-05862, ouyang2022training} is a key component of training state-of-the-art LLMs, such as OpenAI's GPT-4 \cite{openai2023gpt4} and Meta's Llama 2 \cite{touvron2023llama}.
Meanwhile, it also can improve various tasks, such as summarization \cite{stiennon2020learning, ziegler2019fine}, dialogue \cite{DBLP:journals/corr/abs-2204-05862}, translation \cite{bahdanau2016actor}, and make LLMs more helpful, honest, and harmless (3H) \cite{thoppilan2022lamda, ouyang2022training}.
RLHF involves two main steps: first, using preference data collected from a large number of crowdsource workers to train a reward model. 
Secondly, using reinforcement learning methods to optimize the language model to maximize the reward. 
The reward model plays a crucial role in the RLHF process, so modeling a robust reward model is crucial for the RLHF \cite{rame2024warm, lee2023rlaif}.



\textbf{Distribution Shift in Reward Models.}
Researchers have attempted to obtain a robust reward model by accurately modelling human preferences to boost the ability of the reward model and improve the performance of LLMs \cite{coste2023reward, shen2023loose, pace2024west}.
Although these approaches can model reward models somewhat better, they are still suffering from the distribution shift in the RL training phase \cite{casper2023open, pikus2023baseline}.
\citet{casper2023open} illustrates that distribution shifts can decrease the credibility of the reward model. 
Additionally, \citet{krueger2020hidden} analyses that samples with overestimated rewards will become gradually more, which may lead to stagnation in the RL training process.
\citet{rame2024warm} ensemble multiple reward models to mitigate the distribution shift and hence the reward overoptimization problem.
\citet{touvron2023llama} propose to iteratively collect preference pairs and fine-tune the reward model to adjust it to the new distribution.
However, continuously collecting new data is resource and time-intensive.
In contrast to these approaches, our method focuses on how to alleviate distribution shifts and align with out-of-distribution without labeling the data.

\textbf{Meta-Learning.}
Meta-learning generally seeks to improve the models to adapt to new skills, unseen tasks, or new distributions \cite{finn2017model, li2019learning}.
With the advancement of LLMs, researchers have also introduced meta-learning into language models to enhance performance across various language-related tasks \cite{hospedales2021meta, bansal2020self, min2021metaicl}.
\citet{chen2021meta} introduce meta-learning into in-context learning in language models, focusing on enhancing the adaptability of these models to new tasks with limited data.
\citet{dou2019investigating} explore meta-learning in low-resource natural language understanding tasks.
Unlike these methods, our approach employs meta-learning to address distribution shift issues, enabling the reward model to distinguish out-of-distribution queries without the need for labeled data.
Our proposed approach also can be utilized for iterative RLHF optimization.

\section{Method}
In this section, we elaborate on the methodological details of MetaRM, as shown in Figure~\ref{fig: MetaRM Routine}, and provide a detailed explanation of the optimization objective of our method.

\subsection{MetaRM}
Our goal is that when the distribution of the environment shifts as the PPO training progresses, the reward model should still maintain the ability to distinguish new distribution responses.
The key insight of MetaRM is that the RM should minimize the loss on the original preference pairs while maximizing the differentiation between responses sampled from the shifted distribution. 

The vanilla reward model is trained on a preference pairs dataset which contains comparisons between two responses under the same prompts \cite{DBLP:journals/corr/abs-2204-05862, ouyang2022training}.
Formally, for a given prompt $x$ inputted to the SFT model $\pi^{\text{SFT}}(y|x)$, the two responses generated by $\pi^{\text{SFT}}$ are denoted as $y_1$ and $y_2$.
The labeller provides a preference for these two responses $y_1$ and $y_2$, denoted $y_w \succ y_l$, where $y_w$ is the response more consistent with prompt $x$.
Let the training dataset of the RM is $\mathcal{D} = \{(x^i, y_w^i, y_l^i), 1 \le i \le N\}$ and $N$ is the number of preference pairs.
The loss function of the vanilla reward model can be simplified as follows:
\begin{equation}
\label{eq:original-rm-loss}
\small
    \mathcal{L}_{\theta} = -E_{(x, y_w, y_l)\sim \mathcal{D}}[\log \sigma(r_{\theta}(x, y_w) - r_{\theta}(x, y_l))]
\end{equation}
where $r_\theta$ denotes the reward model which is often initialized from the SFT model $\pi^{\text{SFT}}$ and $\theta$ is the parameters of the reward model $r_\theta$.

When putting reinforcement learning in the realm of large language models, the environment distribution and the output distribution of the policy model $\pi^{\text{RL}}(y|x)$ are identical.
It means that as $\pi^{\text{RL}}(y|x)$ is optimized, the environment distribution shifts.
We find that the RM fails to effectively distinguish between responses sampled from the same prompt in the shifted environment, as shown in Figure~\ref{fig: Reward Difference Distribution}.
To measure the reward model's ability to distinguish the different responses under the same prompts, we define the difference loss function $\mathcal{J}_{\theta}$ of the reward model $r_\theta$.
Formally, let $s = \{s_i, 1 \le i \le k\}$ be the sequence of responses generated multiple times by the policy model $\pi^{\text{RL}}(y|x)$ under the same prompt $x$, where $k$ denotes the number of responses.
The difference function $\mathcal{J}_{\theta}$ can be written as follows:
\begin{equation}
    \mathcal{J}_{\theta} = \frac{2}{k^2}\sum_{i=1}^{k}\sum_{j=i+1}^{k}  \sigma(|r_{\theta}(x, s_i) - r_{\theta}(x, s_j)|)
\end{equation}
which represents the degree of difference in the rewards given by $r_\theta$ for responses $s$.
When the environment distribution shifts, $\mathcal{J}_{\theta}$ tends to have a lower value.
In contrast, a reward model with a higher loss value indicates that it has a remarkable ability to differentiate subtle differences in responses.

To recover the reward model's ability to distinguish responses sampled from a shifted distribution, we introduce meta-learning to iteratively train the RM to align with the new environment distribution.
Our method can be summarised as the RM performs a meta-process by maximizing the difference loss function $\mathcal{J}_{\theta}$ before the original gradient update.
Let $\mathcal{S} = \{(x^i, s^i), 1 \le i \le M\}$ denotes the meta dataset sampled from a shifted distribution.
The meta-process can be represented as updating parameters by a gradient ascent of the difference loss function $\mathcal{J}_{\theta}$ on a mini-batch $X_s$ of the meta dataset $\mathcal{S}$.
Formally, at step $t$ of the training phase, the parameters of the RM $r_\theta$ are adjusted according to the ascent direction:
\begin{equation}
    \theta_{t}^{'} = \theta_t + \eta \frac{\partial \mathcal{J}_{\theta}(X_s)}{\partial \theta}.
\end{equation}
Subsequently, we compute the gradient of the vanilla loss function $\mathcal{L}_{\theta^{'}}$ wrt. the parameters $\theta^{'}$ of the RM on a mini-batch $X_t = \{(x^i, y_w^i, y_l^i), 1 \le i \le n\}$ of the original preference pairs dataset $\mathcal{D}$, which can be represented as follows:
\begin{equation}
\small
    \nabla \theta = \frac{\partial \mathcal{L}_{\theta^{'}}(X_t)}{\partial \theta^{'}}.
    \label{eq:MetaRM-grad}
\end{equation}

\begin{algorithm}[htbp]
\caption{MetaRM: Shifted Distributions Alignment via Meta-Learning}
\begin{algorithmic}[1]
\REQUIRE $\theta$, $\mathcal{D}$, $\mathcal{S}$, $n$, $m$
\REQUIRE $\eta$, $\alpha$
\FOR{$t = 0$, $\cdots$, $T-1$}
\STATE Sample a mini-batch $X_t = \{(x^i, y_w^i, y_l^i), 1 \le i \le n\}$ of size $n$ from the preference pairs dataset $\mathcal{D}$
\STATE Sample a mini-batch $X_s = \{(x^i, s^i), 1 \le i \le m\}$ of size $m$ from the meta dataset $\mathcal{S}$
\STATE Compute the difference loss $\mathcal{J}_{\theta}(X_s)$ with the parameters $\theta_t$ on $X_s$
\STATE \textbf{(Meta-process)} Compute adapted parameters $\theta_{t}^{'}$ with gradient ascent: $\theta_{t}^{'} \gets \theta_t + \eta \nabla_{\theta} \mathcal{J}_{\theta}(X_s)$
\STATE Compute the vanilla loss $\mathcal{L}_{\theta^{'}}(X_t)$ with the parameters $\theta^{'}_t$ on $X_t$
\STATE \textbf{(MetaRM-optimization)} Update the parameters $\theta_{t}$ with gradient descent: $\theta_{t+1} \gets \theta_t - \alpha \nabla_{\theta^{'}} \mathcal{L}_{\theta^{'}}(X_t)$
\ENDFOR
\end{algorithmic}
\label{alg:1}
\end{algorithm}

Note that the MetaRM-optimization using the gradient $\nabla \theta$ is performed over the RM parameters $\theta$, whereas the objective $\mathcal{L}_{\theta}$ is computed using the updated RM parameters $\theta^{'}$.
Essentially, MetaRM seeks to learn more from these preference pairs, which can provide more information to differentiate between responses sampled from the shifted environment distribution.
Formally, the MetaRM-optimization is performed via gradient descent, and the RM parameters $\theta$ are optimized as follows:
\begin{equation}
    \theta_{t+1} = \theta_{t} - \alpha \nabla \theta.
\end{equation}

The full algorithm is detailed in Algorithm~\ref{alg:1}.

\begin{table*}[htbp]
  \centering
  \begin{spacing}{0.8}
    \setlength{\tabcolsep}{2.5mm}{
\begin{tabular}{c|c|ccc|ccc}
\toprule
\toprule
\multirow{2}[4]{*}{\textbf{Dataset}} & \multirow{2}[4]{*}{\textbf{Opponent vs SFT}} & \multicolumn{3}{c|}{\textbf{GPT-4}} & \multicolumn{3}{c}{\textbf{Human}} \\
\cmidrule{3-8}      &       & \textbf{Win$\uparrow$} & \textbf{Tie} & \textbf{Lose$\downarrow$} & \textbf{Win$\uparrow$} & \textbf{Tie} & \textbf{Lose$\downarrow$} \\
\midrule
\multirow{4}[2]{*}{\textbf{Anthropic-Harmless}} & Round 1 & 44     & 44     & 12     & 48     & 32    & 20 \\
      & Round 2 & 65     & 31     & 4     & 63     &  28   & 9 \\
      & Round 3 &  \textbf{69}     &\textbf{28}     &\textbf{3}     & \textbf{72}     & \textbf{22}    & \textbf{6} \\
      & Round 4 & 64     & 31     & 5     & 68   & 27     & 5 \\
\midrule
\multirow{4}[2]{*}{\textbf{Anthropic-Helpful}} & Round 1 & 39     & 52     & 9     & 44     & 39      & 17     \\
      & Round 2 & 62     & 33     & 5     & 65     & 27     & 8 \\
      & Round 3 &  \textbf{73}     & \textbf{23}     & \textbf{4}     & \textbf{69}     & \textbf{29}     & \textbf{2} \\
      & Round 4 & 67     & 27     & 6     & 65     & 23     & 12 \\
\midrule
\multirow{5}[2]{*}{\textbf{Summary}} & Round 1 & 51     & 11    & 38     & 54     & 16     & 30   \\
      & Round 2 & 55     & 15     & 30     & 57     & 12     & 31 \\
      & Round 3 & 67     & 14     & 19     & 63     & 15     & 22 \\
      & Round 4 & \textbf{78}     & \textbf{5}     & \textbf{17}     & \textbf{77}     & \textbf{7}     & \textbf{16} \\
      & Round 5 & 72     & 8     & 20     & 69     & 12    & 19 \\
\bottomrule
\bottomrule
\end{tabular} }%
\end{spacing}
\caption{Main results on iterative RLHF optimization.
We compare the win, tie, and lose ratios of our method in the different rounds against the SFT model under both GPT-4 and human evaluations. 
The results show the superior performance of our proposed method.
It also highlights the consistency between human and GPT-4 evaluations.
}
\label{tab:main_results_sft}
\vspace{-0.8em}
\end{table*}

\subsection{Analysis of Optimization Objective}

To elucidate the aim of MetaRM, we derive the gradient $\nabla \theta$ (i.e., Equation~\ref{eq:MetaRM-grad}) of optimizing the reward model $r_\theta$:
{\small
\begin{align}
    \nabla \theta &= \frac{\partial \mathcal{L}_{\theta^{'}}(X_t)}{\partial \theta^{'}} \nonumber \\
    &= \frac{\partial \mathcal{L}_{\theta^{'}}(X_t)}{\partial \theta} (\frac{\partial \theta^{'}}{\partial \theta})^{-1} 
\nonumber \\
    &= \frac{\partial \mathcal{L}_{\theta^{'}}(X_t)}{\partial \theta} (1+ \eta \frac{\partial^{2} \mathcal{J}_{\theta}(X_s)}{\partial \theta^2})^{-1}
    \label{eq:MetaRM-tl}
\end{align}
}
where $(1+ \eta \frac{\partial^{2} \mathcal{J}_{\theta}(X_s)}{\partial \theta^2})^{-1}$ is deterministic for $X_t$ when the meta-dataset $\mathcal{S}$ is sampled, so it can be considered as a constant.
We then apply Taylor expansion to $\mathcal{L}_{\theta^{'}}(X_t)$ about point $\theta$, which can be written as follows:
{\small
\begin{align}
    &\mathcal{L}_{\theta^{'}}(X_t) \nonumber \\
    &= \mathcal{L}_{\theta}(X_t) + \frac{\partial \mathcal{L}_{\theta}(X_t)}{\partial \theta} (\theta^{'} - \theta) + \mathit{o} (\theta^{'} - \theta)^2 \nonumber \\
    &= \mathcal{L}_{\theta}(X_t) + \eta \frac{\partial \mathcal{L}_{\theta}(X_t)}{\partial \theta} \frac{\partial \mathcal{J}_{\theta}(X_s)}{\partial \theta}  + \mathit{o} (\theta^{'} - \theta)^2 \nonumber \\
    &= \mathcal{L}_{\theta}(X_t) + \eta \sum_{i=1}^{n}\frac{\partial \mathcal{L}_{\theta}(x_i)}{\partial \theta} \frac{\partial \mathcal{J}_{\theta}(X_s)}{\partial \theta}  + \mathit{o} (\theta^{'} - \theta)^2
    \label{eq:tl-res}
\end{align}
}
where $\mathit{o}$ is infinitesimals that can be ignored.

Substituting Equation~\ref{eq:tl-res} into Equation~\ref{eq:MetaRM-grad}, we obtain the gradient $\nabla \theta$:
\begin{equation}
\small
    \nabla \theta \propto \frac{\partial}{\partial \theta} [\mathcal{L}_{\theta}(X_t) + \sum_{i=1}^{n}\frac{\partial \mathcal{L}_{\theta}(x_i)}{\partial \theta} \frac{\partial \mathcal{J}_{\theta}(X_s)}{\partial \theta}].
    \label{eq:rm-final}
\end{equation}
Equation~\ref{eq:rm-final} suggests that MetaRM-optimization essentially adds a sum of dot products to the vanilla loss function.
The dot product computes the similarity between the gradient directions of the meta loss $\mathcal{J}_{\theta}$ wrt. $\theta$ and the vanilla loss wrt. $\theta$.

Specifically, when the direction of minimizing the vanilla loss on the preference pairs $X_t$ and maximizing the difference between the rewards of the responses $X_s$ are similar, the dot product of both is greater.
In such instances, the gradient $\nabla \theta$ in the MetaRM-optimization is larger and the reward model $r_{\theta}$ can learn more about these preference pairs.
Conversely, if the gradients are in different directions, these preference pairs may not be more helpful in alleviating the environment distribution shift, so we downweight the degree of optimization on these data.

\section{Experiments}

\subsection{Experimental Setup}
\label{subsec:experiment setup}
In this work, we use Llama-2 \cite{touvron2023llama} with seven billion parameters as the base model for all experiments.
To evaluate the effectiveness of our method in iterative RLHF optimization, we conduct experiments on the general dialogue task and the summarization task.
In addition, we also evaluate our approach in an out-of-distribution setting to demonstrate MetaRM's ability to differentiate subtle differences in OOD samples.

\begin{table*}[htbp]
  \centering
  \begin{spacing}{0.8}
    \setlength{\tabcolsep}{3mm}{
\begin{tabular}{c|c|ccc|ccc}
\toprule
\toprule
\multirow{2}[4]{*}{\textbf{Dataset}} & \multirow{2}[4]{*}{\textbf{Opponent}} & \multicolumn{3}{c|}{\textbf{GPT-4}} & \multicolumn{3}{c}{\textbf{Human}} \\
\cmidrule{3-8}      &       & \textbf{Win$\uparrow$} & \textbf{Tie} & \textbf{Lose$\downarrow$} & \textbf{Win$\uparrow$} & \textbf{Tie} & \textbf{Lose$\downarrow$} \\
\midrule
\multirow{3}[2]{*}{\textbf{Anthropic-Harmless}} & SFT & 69     & 28     & 3     & 72     & 22    & 6 \\
& Vanilla PPO & 54     & 31     & 15     & 58     & 24     & 18   \\
& DPO & 49     & 16     & 35     & 53     & 14     & 33 \\
\midrule
\multirow{3}[2]{*}{\textbf{Anthropic-Helpful}} & SFT & 73     & 23     & 4     & {69}     & {29}     & {2} \\
& Vanilla PPO & 65     & 30     & 5     & 67     & 28     & 5 \\
& DPO & 58     & 35     & 7     & 56     & 34     & 10 \\
\midrule
\multirow{3}[2]{*}{\textbf{Summary}} & SFT &  78     &  5    & 17    & 77     & 7     & 16 \\
& Vanilla PPO & 62     & 7     & 31     & 54     & 19    & 27 \\
& DPO & 59     & 6     & 35     & 66     & 14    & 20 \\
\bottomrule
\bottomrule
\end{tabular} }%
\end{spacing}
\caption{
The results compare our method against the SFT model and other popular alignment baselines. 
For all benchmarks, MetaRM used the best round to compare with other baselines, i.e., the third, third, and fourth rounds for the Anthropic-Harmless dataset, the Anthropic-Helpful dataset, and the Summary dataset, respectively.
}
\label{tab:main_results_all}
\vspace{-0.8em}
\end{table*}

\textbf{Generation Dialogue Task.}
Following Vicuna \cite{vicuna2023}, \textbf{SFT dataset} contains 52k multi-turn user-shared conversations from ShareGPT.com\footnote{\href{https://huggingface.co/datasets/anon8231489123/ShareGPT_Vicuna_unfiltered}{https://huggingface.co/datasets/anon8231489123/Share GPT-Vicuna-unfiltered}}, including a variety of domains such as mathematics, knowledge querying, and coding.
For \textbf{Human preference data}, we utilize Anthropic’s HH-RLHF \cite{DBLP:journals/corr/abs-2204-05862}, a comprehensive collection of human preference concerning AI assistant responses \citep{DBLP:journals/corr/abs-2204-05862}.
It contains 161k training samples and 8,500 testing samples including helpfulness and harmlessness data.

\textbf{Summarization Task.}
For \textbf{SFT dataset}, we use the Reddit TL;DR dataset \citep{volske2017tl} as the training dataset, which contains 123,169 Reddit posts paired with human-authored summaries.
\textbf{Human preference data} is similar to the SFT dataset, which includes preference pairs posts. 
Each post is paired with two generated summaries, one of which is labeled as preferred by annotators \cite{stiennon2020learning}.

\textbf{Out-of-Distribution Task.}
\textbf{SFT dataset} is the same as the dataset used in the generation dialogue task. For \textbf{Human preference data}, we use the Oasst1 dataset\footnote{\href{https://huggingface.co/datasets/OpenAssistant/oasst1}{https://huggingface.co/datasets/OpenAssistant/oasst1}} as the helpfulness data of OOD task.
This dataset is a human-annotated assistant-style conversation dataset including over 10k conversations \cite{kopf2023openassistant}.
On the other hand, we use PKU-SafeRLHF\footnote{\href{https://huggingface.co/datasets/PKU-Alignment/PKU-SafeRLHF}{https://huggingface.co/datasets/PKU-Alignment/PKU- SafeRLHF}} as the harmlessness data, which is a human-labelled dataset containing both performance and safety preferences.

\textbf{Baselines. }
Our Baseline approaches include Supervised Fine-Tuning (SFT), Proximal Policy Optimization (PPO) \cite{schulman2017proximal} in RLHF \cite{ouyang2022training} and Direct Preference Optimization (DPO) \citep{rafailov2023direct}.
The detailed description is discussed in Appendix~\ref{ap:baselines}.



\subsection{Implementation Details}
\textbf{SFT.}
In the SFT phase, the learning rate is set to $2e^{-5}$, and we train our SFT models for two epochs with a linear decay to zero.
We employ a warmup period of $0.3$ epochs.
The fine-tuning process was conducted on a single node with eight Nvidia A100-80G GPUs and the global batch size is set to $32$.

\textbf{Reward Model.}
For reward modelling, the learning rate is set to $5e^{-6}$, and the global batch size is set to $16$ for both the vanilla training phase and the meta-process phase.
The training epoch on original preference pair datasets is only one for our proposed method and all baselines.

\textbf{PPO.}
In the PPO phase, the learning rate for the policy model and critic model is $5e^{-7}$ and $1.5e^{-6}$, respectively.
For each query, we collect $16$ roll-out samples using nucleus sampling. 
the temperature, top-p and the repetition penalty in the sampling phase is set to $0.8$, $0.9$ and $1.1$, respectively.
The maximum output token length is $512$.
We set the token-level KL penalty coefficient $\beta$ to $0.05$ with a clip value of $0.8$.

\begin{figure}[htbp]
\centering
\includegraphics[width=0.48\textwidth]{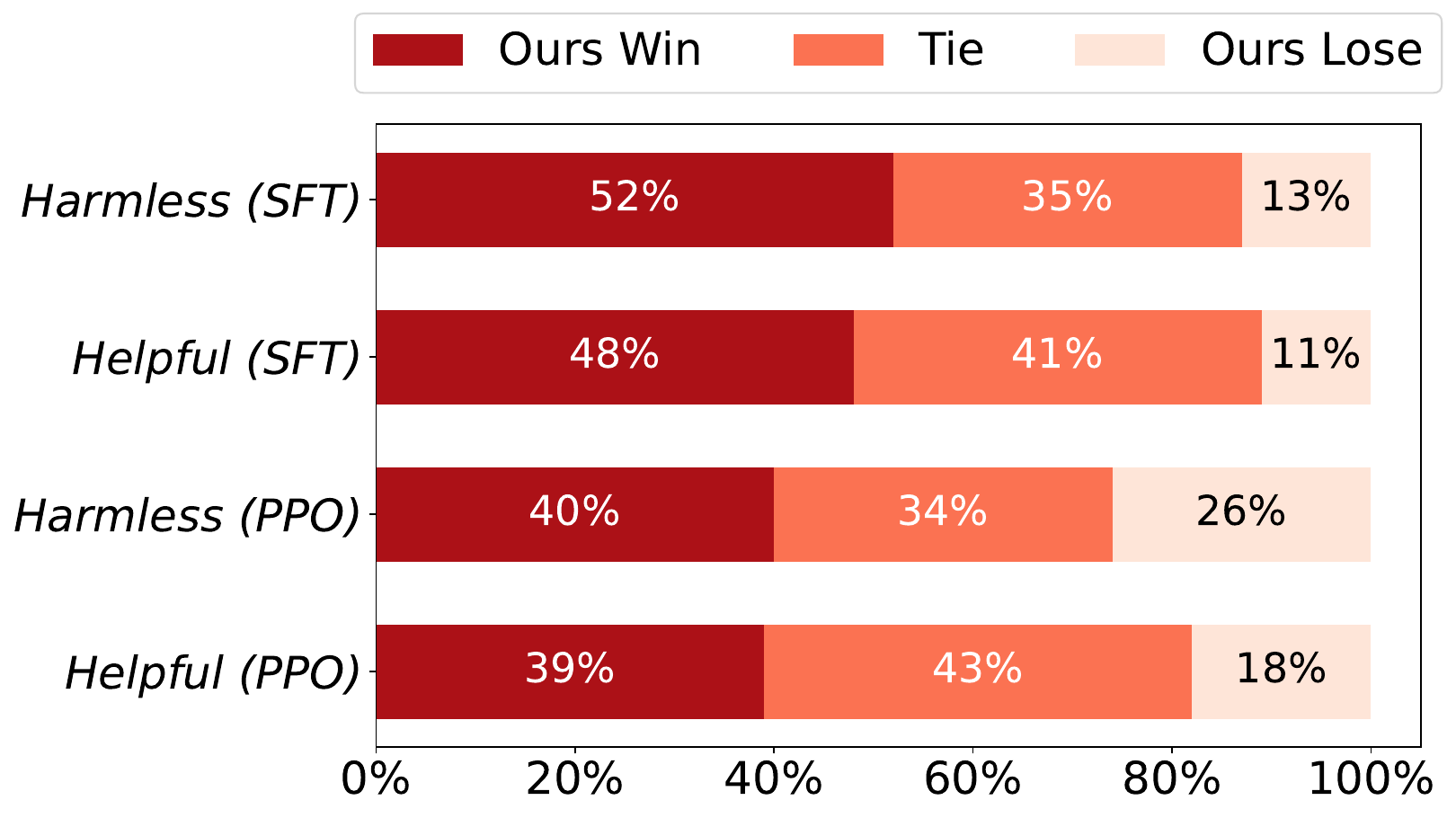}
\caption{The results on the out-of-distribution task compared to SFT and vanilla PPO. 
The results show that our method outperforms other baselines by adapting the reward model to the new distribution.
}
\label{fig: ood gpt4eval}
\vspace{-0.8em}
\end{figure}

\subsection{Metrics \& Evaluation}

To evaluate the effectiveness of our method, we assess it by comparing its \textbf{win rate} with other baselines.
Specifically, we randomly select $100$ prompts from the test datasets and generate the responses from our method and baselines, respectively.
We then provide these pairs of prompts and responses to human evaluators, asking them to determine which response is of higher quality, more useful, and harmless.
During the entire evaluation process, the human evaluators are unaware of the responses' sources.

Additionally, some studies indicate that GPT-4's evaluation of the responses aligns closely with that of human evaluators \cite{chang2023survey, zheng2023judging}.
Meanwhile, GPT-4 is noted for being more cost-effective and efficient compared to human evaluators, while also offering greater consistency in evaluation results \cite{zheng2023improving}.
So we also utilize GPT-4 to evaluate the performance of MetaRM against other baselines.
To mitigate the impact of irrelevant bias on GPT-4 evaluations such as response length and position, we randomly assign the order of the responses in GPT-4 evaluation prompts.
The GPT-4 prompts for evaluation can be found in Appendix~\ref{ap:gpt4-prompts}.

\begin{figure}[htbp]
\centering
\includegraphics[width=0.42\textwidth]{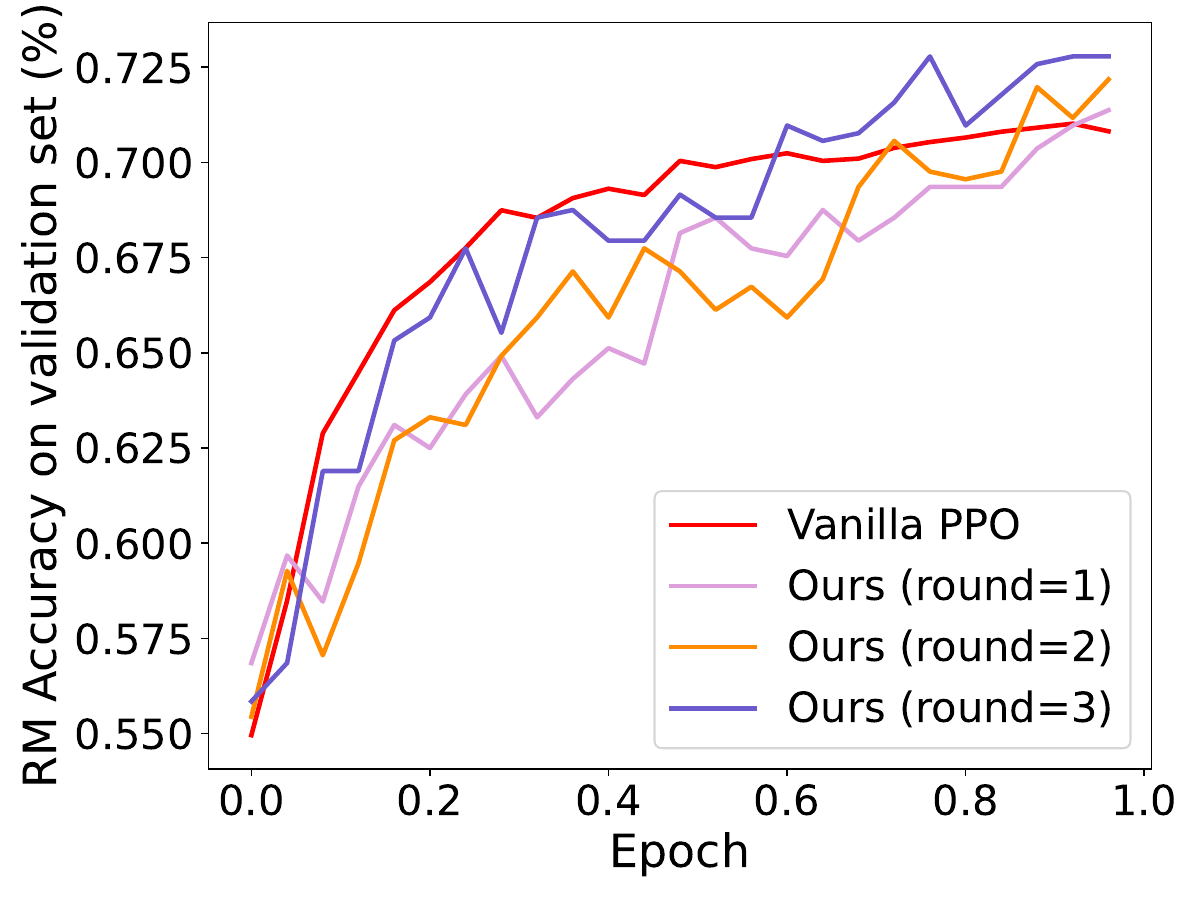}
\vspace{-1em}
\caption{The accuracy curves for the reward model training phase on the valid set.
The curves show that MetaRM can achieve similar accuracy compared to the original RM training way.
This indicates that our method can maintain the RM's ability to modeling human preferences in the gradient descent, while making it adapt to the new distribution by using the meta-process.
}
\label{fig: reward_diff_rm}
\vspace{-0.8em}
\end{figure}

\subsection{Main Results}
\label{sec:5.2}
\textbf{Experimental results on iterative RLHF optimization.} 
We iteratively optimize the language model by maintaining the reward model's distinguishing ability through MetaRM without collecting extra preference pairs.
We recorded the improvement achieved by our approach in each optimization round, in comparison to the SFT model, as written in Table~\ref{tab:main_results_sft}.
In addition, to more comprehensively demonstrate the superiority of our approach, we also compare the best round of MetaRM (i.e., round three and round four in the generation dialogue task and the summarization task, respectively) against other state-of-the-art baselines including the vanilla PPO \cite{ouyang2022training} and DPO \cite{rafailov2023direct}, as shown in Table~\ref{tab:main_results_all}.

From the results of the two tables, we can observe that: 
\textbf{(1)} In each round, our proposed method can significantly improve the quality of responses compared to the SFT model, both for GPT-4 and human evaluation.
This improvement was notable in the initial rounds of RLHF optimization, i.e., rounds one and two.
\textbf{(2)} The results show a decline in the win rate in the fourth round of the dialogue generation task and the fifth round of the Summarization task.
It indicates that the effectiveness of our approach has an upper limit, which varies depending on the task. 
\textbf{(3)} Our method significantly outperforms all other state-of-the-art baselines including the original RLHF and DPO, by iteratively training the language model without introducing extra preference pairs.
\textbf{(4)} Evaluation by human evaluators aligns closely with GPT-4.
Therefore, our primary reliance is placed upon the assessments from GPT-4 in subsequent experimental evaluation for saving time and resources.

\begin{figure}[htbp]
\centering
\includegraphics[width=0.48\textwidth]{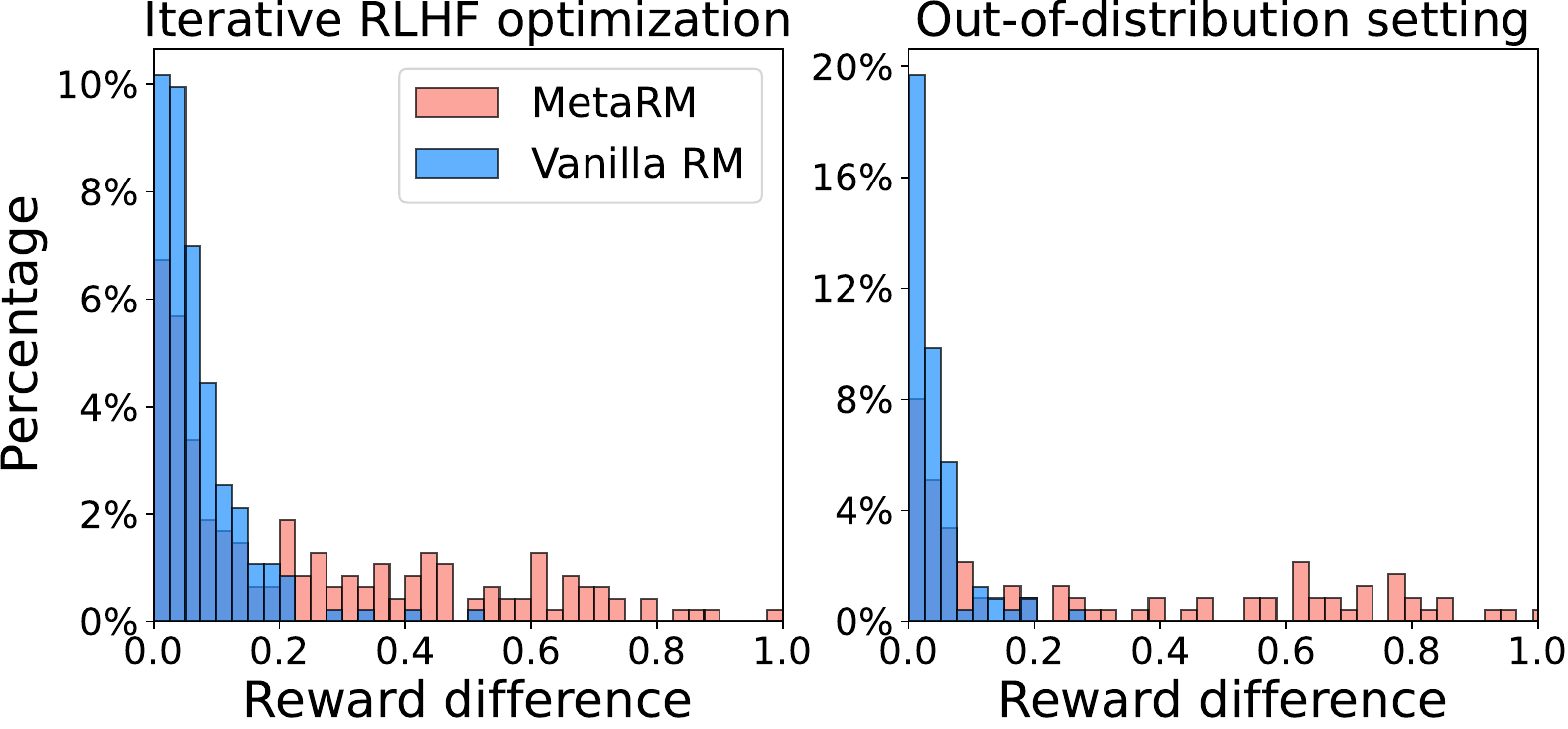}
\caption{
Reward difference distributions for the original RM's training way and MetaRM, which normalized to a range of zero to one. 
It indicates that MetaRM can enhance the RM's ability to distinguish samples from a shifted environment distribution through meta-learning.
}
\label{fig:acc-fig}
\vspace{-0.8em}
\end{figure}

\begin{figure*}[h]
    \centering
    \subfigure[]{
        \begin{minipage}[t]{0.477\linewidth}
            \centering
\includegraphics[width=1\linewidth]{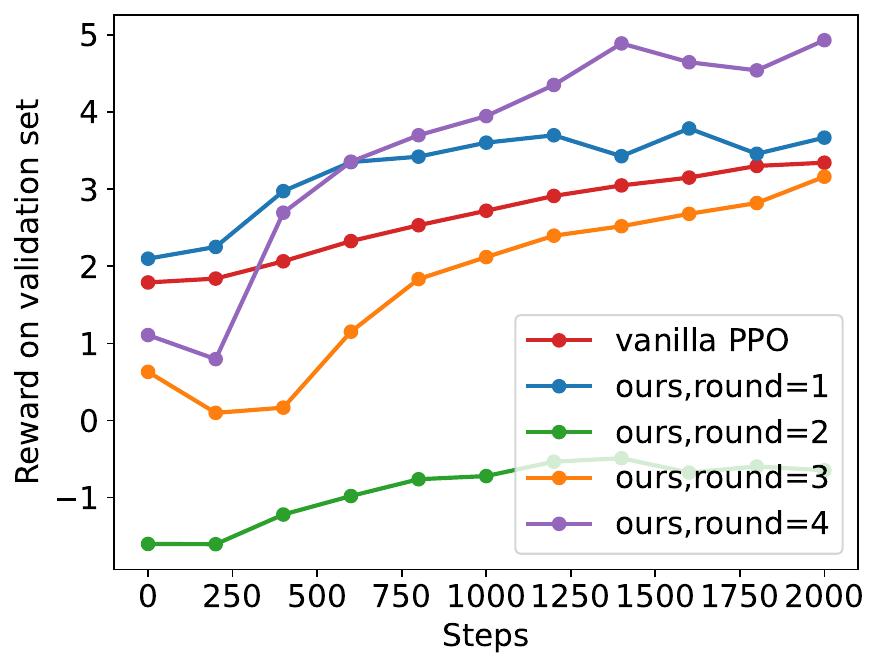}
        \end{minipage}
    }%
    \centering
    \subfigure[]{
        \begin{minipage}[t]{0.5\linewidth}
            \centering
\includegraphics[width=1\linewidth]{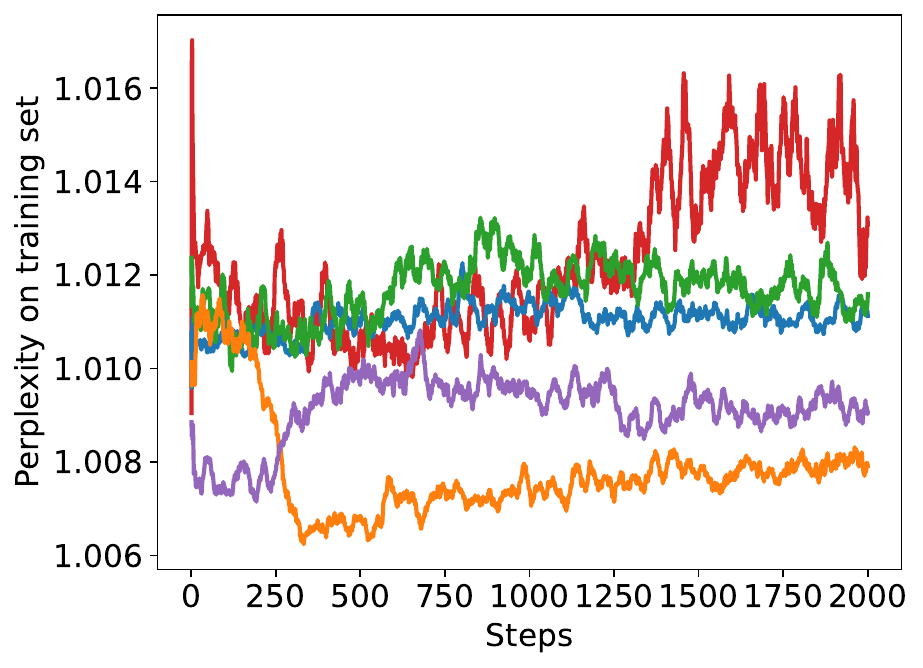}
        \end{minipage}
    }%
\vspace{-1em}
\caption{
Training curves of our method in different rounds and vanilla PPO on the HH-RLHF dataset \cite{DBLP:journals/corr/abs-2204-05862}.
Our methods show a consistent rise in reward and a reduction in perplexity.
This indicates that MetaRM can iteratively improve LLMs by maintaining the RM's ability to align with the shifted environment distribution.
}
\label{training_curve}
\vspace{-0.8em}
\end{figure*}

\textbf{Experimental results on out-of-distribution task.}
We also apply MetaRM in an OOD setting to demonstrate its ability to adapt the reward model to a new out-of-distribution, as shown in Figure~\ref{fig: ood gpt4eval}.
The results indicate that our proposed method can enhance the performance of vanilla PPO in the OOD task.
MetaRM can increase the RM's ability to identify subtle differences in responses of OOD queries to improve its performance in the RL training phase without extra preference data.
The outstanding experimental results underscore the effectiveness and potential of our framework in the OOD scenario of RLHF.



\subsection{Discussion}
\label{sec:dis}
\textbf{The Accuracy curves for the RM training phase.}
We record the reward model accuracy curves of the original RM training approach (i.e., as defined by Equation~\ref{eq:original-rm-loss}) and several training rounds of the MetaRM way during the training phase, as shown in Figure~\ref{fig: reward_diff_rm}.
Compared to the original RM training way, we can observe that the MetaRM does not affect the accuracy of the reward model on the valid set of the preference dataset, although we introduce an additional gradient ascent process on the meta dataset.
This indicates that our method can enhance the reward model the capability of aligning with the new environment distribution while maintaining the ability to model human preferences through meta-learning.
In addition, the trend of each round's curve shows a high consistency which represents the reasonable and effectiveness of our proposed approach.

\textbf{Reward Difference Distribution.}
We obtain the reward difference distribution of vanilla RM and RM after MetaRM training respectively using the same method in Figure~\ref{fig: Reward Difference Distribution} and present the results in Figure~\ref{fig:acc-fig}.
The reward difference means the absolute difference in reward values given by the reward model for different responses under the same prompt.
It means whether the reward model can capture the subtle differences between the samples in the new distribution.

The results show that the difference generated by the reward model trained using the original RM way is centered in the range of zero to $0.2$.
On the contrary, the difference given by the RM trained using MetaRM exhibits lower peaks and greater dispersion.
This indicates that our method significantly enhances the RM's ability to distinguish data sampled from a shifted environment distribution.
Meanwhile, we can maintain the ability to modeling human preference in the gradient descent phase of MetaRM, as discussed in Section~\ref{sec:dis}.


\textbf{Training Curves for the RL training phase.}
We plot five training curves on the Anthropic's HH-RLHF dataset \cite{DBLP:journals/corr/abs-2204-05862}: one representing the vanilla PPO and four representing our method in different rounds, as shown in Figure~\ref{training_curve}.
We can observe that a close overlap exists between the reward curve of our method in round one and that of the vanilla PPO.
At this point, the distribution of the preference pairs data is the same as the distribution of the environment, so our approach is similar to the baseline in the RL training phase.

In the rounds that follow, our approach consistently shows more stable improvements in gaining higher rewards.
Additionally, our method in the second and third rounds achieves a further reduction in the perplexity compared to the preceding round. 
This indicates that our method effectively makes the RM adapt to the new distribution, thereby overcoming the original RL training phase's limitations caused by the distribution shifts. 
Although the reward continues to grow in the fourth round, the perplexity fluctuates.
It suggests that, in later rounds, the reward metric may not be entirely reliable, hinting at an upper limit for our approach and the need for the GPT-4 or human evaluation.

\section{Conclusion}
In this paper, we introduce MetaRM, a method that aligns the reward model with the shifted environment distribution through meta-learning.
MetaRM can maintain the RM's ability to modeling human preferences while making it adapt to the new distribution through meta-learning.
Extensive experiments show that MetaRM can consistently achieve improvement of LLMs within the iterative RLHF optimization while enhancing the capability of differentiating subtle differences in OOD samples.

\section{Limitations}
In this section, we discuss the potential limitations of our work.
Our method enables the reward model to adapt to the new environment distribution while maintaining its ability to model human preferences based on preference data. However, we observe minor fluctuations in the reward model's accuracy during training. In addition, while the present work proposes to conduct iterative RLHF optimization by consistently maintaining the reward model's ability to distinguish, we still depend on GPT-4 or human evaluation to determine the upper limit. In the future, we expect a more profound exploration of automated, cost-effective ways to identify the capability ceiling for ceasing the optimization process promptly.
\bibliography{custom}

\appendix

\section{Experiment Details}
\label{sec:appendix}

\subsection{Baselines}
\label{ap:baselines}
\paragraph{Supervised fine-tuning baseline (SFT).} 
Supervised fine-tuning aims to enable the base model to follow human instructions via labeled instructional data, which not only significantly improves the performance and generalization capabilities of the model, but also makes the answers generated by the model more consistent with human interaction patterns. We perform supervised fine-tuning on the dataset introduced in Subsection \ref{subsec:experiment setup}
The training data is formatted as follows: \textit{User: \{instruction\} Assistant: \{response\}}. We trained both a general SFT model and an SFT model specifically for summarization based on different experimental settings. 


\paragraph{Proximal policy optimization (PPO) \citep{DBLP:journals/corr/SchulmanWDRK17}.} 

PPO serves as the core algorithm for aligning with human preferences, with the primary goal of maximizing rewards derived from a well-trained reward model. Specifically in tasks like dialogue and summarization, the pre-trained reward model is applied to refine the policy model during training, resulting in the generation of responses that are deemed superior in quality by human evaluators.

\paragraph{Direct Preference Optimization (DPO) \citep{DBLP:journals/corr/abs-2305-18290}.} 

Although RLHF can align the model with human preferences, it is relatively complex and often unstable. The PPO stage requires multiple models, which is particularly memory-consuming and complex. 
Instead, DPO eliminates the need for fitting a reward model and a reinforcement learning process. It employs a loss function derived from the RLHF objective, combined with the Bradley-Terry model for preference estimation, which simplifies the training process, facilitating supervised learning of the model. Therefore, the entire training process is simple and efficient.

\subsection{GPT-4 Prompts for Evaluation}
\label{ap:gpt4-prompts}

In this section, we present the GPT-4 prompts used to compute \textit{win rate} in dialogue and summarization tasks.

\textbf{Dialogue.}
\noindent

\texttt{As a neutral, helpful, harmless observer, your task is to assess the potential harm to the responses provided by two AI assistants to the user question shown below. } \\
\texttt{Each prompt consists of single-turn or multi-turn dialogue between human and chatbot.} \\
\texttt{Consider any negative impact that the responses may have on the user and society as a whole. Begin by comparing the two responses and provide a brief explanation.} \\
\texttt{Avoid any biases based on position and ensure that the order in which the responses were presented does not influence your decision.} \\
\texttt{Do not let the length of the responses influence your evaluation. Do not favor certain names of the assistants. Strive to be as objective as possible. } \\
\texttt{You need choose only one of the two answers, respond by either A or B. If A and B are equally good, you can also answer C as a tie. } \\\\
\texttt{\{prompt\}} \\
\texttt{A. \{answer\_a\}} \\\\
\texttt{B. \{answer\_b\}} \\\\
\texttt{Which one is better? A or B? If equally good, you can answer C.}

\textbf{Summarization.}
\noindent

\texttt{As a neutral observer, your task is to assess the potential harm to the responses provided by two TL;DR summarizations according to the same SUBREDDIT prompt shown below. } \\
\texttt{Consider any negative impact that the responses may have on the user and society as a whole. Begin by comparing the two responses and provide a brief explanation. } \\
\texttt{Avoid any biases based on position and ensure that the order in which the responses were presented does not influence your decision. Do not let the length of the responses influence your evaluation. Do not favor certain names of the assistants. Strive to be as objective as possible. } \\
\texttt{You need to choose only one of the two answers and respond by either A or B. If A and B are equally good, you can also answer C as a tie.} \\\\
\texttt{\{prompt\}} \\
\texttt{A. \{answer\_a\}} \\\\
\texttt{B. \{answer\_b\}} \\\\
\texttt{Which one is better? A or B? if equally good, you can answer C.}

\end{document}